# CLASSIFICATION OF ARTIFICIAL INTELLIGENCE IDS FOR SMURF ATTACK

N.Ugtakhbayar, D.Battulga and Sh.Sodbileg

Department of Communication technology, School of Information Technology,

National University of Mongolia, Mongolia

sdblg@num.edu.mn

*ABSTRACT*

*Many methods have been developed to secure the network infrastructure and communication over the Internet. Intrusion detection is a relatively new addition to such techniques. Intrusion detection systems (IDS) are used to find out if someone has intrusion into or is trying to get it the network. One big problem is amount of Intrusion which is increasing day by day. We need to know about network attack information using IDS, then analysing the effect. Due to the nature of IDSs which are solely signature based, every new intrusion cannot be detected; so it is important to introduce artificial intelligence (AI) methods / techniques in IDS. Introduction of AI necessitates the importance of normalization in intrusions. This work is focused on classification of AI based IDS techniques which will help better design intrusion detection systems in the future. We have also proposed a support vector machine for IDS to detect Smurf attack with much reliable accuracy.*

## 1. INTRODUCTION

One of the problems for computer network's intrusion detection system that passively monitor a network link is the ability of a skilled attacker to evade detection by exploiting ambiguities in the traffic stream as seen by the network intrusion detection system (NIDS). [1] [3] The NIDS may less capability complete analysis for the full range of behaviour allowed by a particular protocol. For instance, attacker may be evade a NIDS that fails to reassemble IP fragments by their attack traffic.
The traffic normalization mechanisms belong to the wide set of tools that help the allocation, control of resources in TCP/IP networks, and base system of improve the reliability NIDS. [3] Many technique can be to classify network traffic. The more skilled method is neural network. However, one requirement of such system is normal traffic. Then a key problem is how to choose the attributes of the input training data. In this paper, we use a support vector machine algorithm based on the KDDCUP 99 dataset. This paper produces a support vector machine for network intrusion detection that can detect smurf attack in network with accuracy of 99.6%. Also this system can predict the percentage of smurf infection in the network with absolute error average from 0% to 5%. One of intrusion detection methods is misuse detection technique. The frequently utilize a rule-based approach. When applied to misuse detection, rules become scenarios for network attacks.

This paper is structured as follows. In Section 2 we discuss related work. Section 3 is the problem statement of intrusion detection system. Section 4 introduces our solution and experiment of intrusion detection with support vector machine. This paper is ended with a conclusion and future work.

## 2. RELATED WORK

During the last years, IDS alert processing techniques have received considerable amount of attention in the research community. [5] Some researcher has proposed machine learning method for IDS alert normalization and classification, in order to reduce the amount of false positives. In 1999, the KDD conference hosted a classifier learning contest, in which the learning task was to build a predictive model to differentiate attacks and normal connections. Contestants trained and tested their classifiers on an intrusion data set provided by MIT labs. All these work above use features of the KDDCUP 99 training and testing data. Sung et al. use SVMs and neural networks to identify important features for 1998 DARPA intrusion detection data. The most related work to ours is done by Dongxi Liu, Chi-Hung Chi and Ming Li. [3] Their achieve goal, they insert a thin layer between the network layer and the link layer. Thus, during every unit time period, only one packet is sent (either a regular packet from the network layer or a padded packet when no regular packet is available, i.e. a dummy packet). Moreover, if the packet to be sent is shorter than an expected length, some extra bytes are padded to its rear so that packets sent by one network node always have constant size with constant rate, i.e., normalized traffic, in which dummy packet and extra bytes together make up the complementary traffic. [3]

## 3. PROBLEM STATEMENT

We need intelligence IDS but the system making first step is traffic classification and normalization. First, packet level normalization would vanish in a puff of logic. However, there is not one implementation, but many and each can behave uniquely under identical stimulation. We are making first, normalization data by using normal traffic. In evaluating normalize; we care about completeness, correctness, and performance [4] [3]. The evaluation presents a challenging problem because by definition most of the functionality of a normalize applies only to unusual or impossible traffic and the results of a normalizer in general are invisible to connection endpoints. The problem here is to obtain test traffic that exhibits the behaviour we wish to normalize; once this is done, we need to ensure that norm correctly normalizes it. TTL setting on incoming packets from attacker causes some packets to be dropped after being inspected by the NIDS. End system only sees packets with longer TTL. Then NIDS misses attack. Fig 1 [11] shows one example of normalization of packets evasion technique.

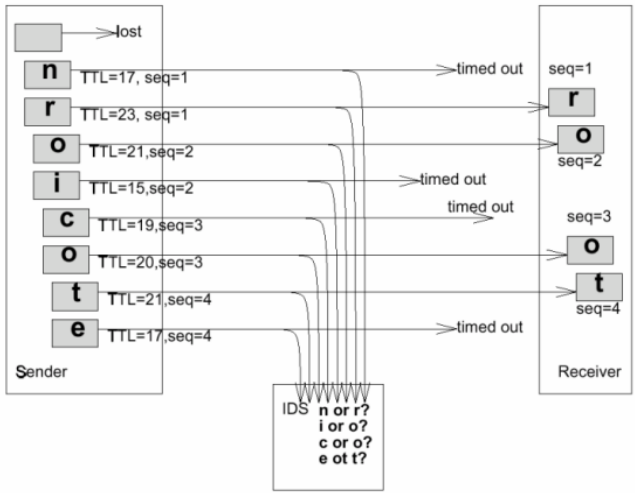

Fig.1. Normalization of packet

**4. SOLUTION**

First, we make IP normalization, because our historical research result shows that IP attack we occurred more frequently on campus network traffic. Two kinds of normalization topology, one is with firewall, shown on Fig 2 and another one is without firewall, shown on Fig 3. In the case of experiment using NIDS without firewall, we can get the log which is marked by alerted, inserted and dropped packets.

Alternative solution is using a host based IDS. But it maybe costly, for used too many hosts in network subnet. Also probable assemble was only intranet network information. [2] [3] [9]

Normalizer can normalize and protect additional functions of firewall. We can get this kind of normalizer from the results which is normalise on IP packets. The normalizer is able to reduce a load of NIDS such as reject packets with wrong checksums.

Also we normalized following parameters of IP packet: Header length, Total packet length, DF flag, MF flag, TTL. The most of attack scan be occurred on structure of IP packet. For instance, TTL problem was occurred, attacker may exploit knowledge of network topology to cause some packets to be dropped. Then this attack type's solution is set all TTL fields to be larger than the longest path inside network. One of more occurred problem is packet with invalid total length field may be dropped or accepted by end system. In most cases NIDS don't know. Solution of this attack, discard if field longer than actual; trim if packet was occurred longer than field normal length.

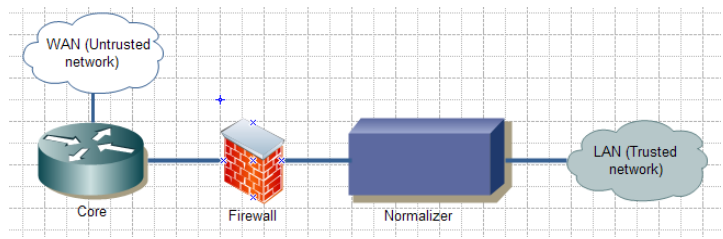

Figure 1 Illustrating possible configurations of a Normalizer /with firewall/

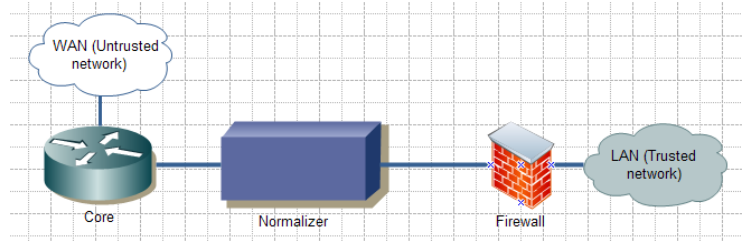

Figure 2 Illustrating possible configurations of a Normalizer /without firewall/

So we are making TCP, ICMP traffic normalization. We are used Linux /kernel 2.5/. One of the problems is network traffic copying from memory to memory. Our experiment result of performance is shown in Table 1. PC can forward 333.000 packets/s. [4]

We are making traffic classification with support vector machine. Then detection making by support vector machine and decision tree algorithm with smurf attack on DTREG tool.

Smurf attack detection condition is hot indication count <=0.0, last 2 second connection counts REJ error >0.21 and <=0.01 and connection to host count <= from 41.2 to 112.3. Shown on Figure 4 is attack normalization example.

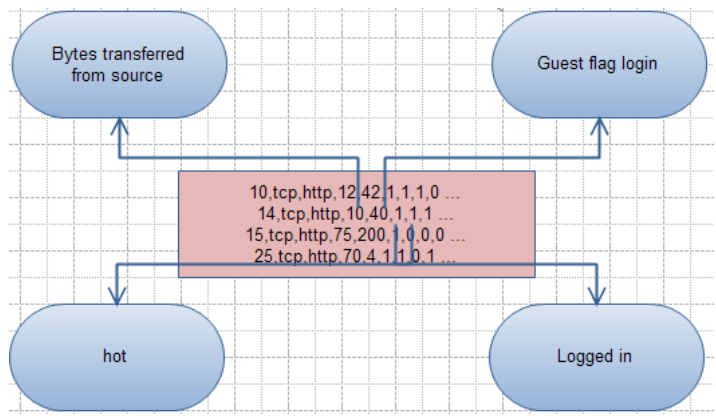

Figure 4. Attack normalization example

This model is used to identify smurf attack from KDDCUP 99 dataset; it can also predict the attack percentage in the network traffic. This model depends only on the data that collected from the IDS and KDDCUP 99. As shown in Figure 5, this model consists of 3 modules: (1) Training data (2) Testing data (3) Cross-over.

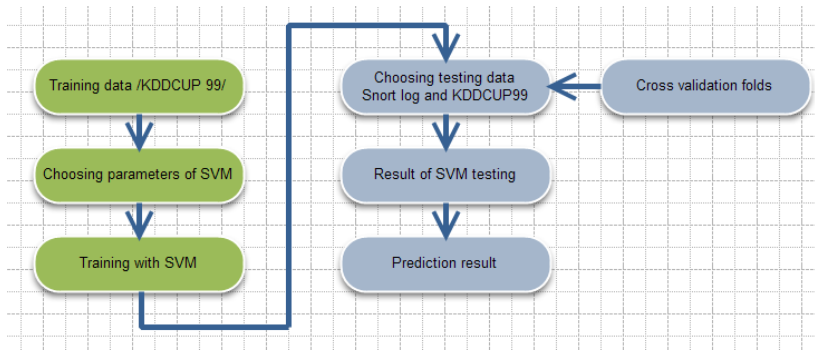

Figure 5. Model of identify smurf attack

In the experiment, we use count of traffic 100.0 and count of parameter 7. Table 1 shows the results of phase identify smurf attack.

Table 1. The result of identify smurf attack.

|  | TP rate | FP rate | Precision | Recall | F-Measure | ROC area | Class |
|---|---|---|---|---|---|---|---|
|  | 0.996 | 0 | 1 | 0.99 | 0.998 | 0.998 | Normal traffic |
|  | 1 | 0.004 | 0.999 | 1 | 0.999 | 0.998 | Smurf attack |
| Weighted Avg. | 0.999 | 0.003 | 0.999 | 0.999 | 0.999 | 0.998 |  |

If it gets less number of parameter which is used to detect the attack from normalized packets, the performance of detection is increase. The method selects parameters which are hot indication count, connection count REJ error, and connection to host count and transferred bytes from source, and detects Smurf attack by 99.9 percent.

## 5. CONCLUSION

In this paper, we proposed a method to detect the occurrence and intensity of Smurf attack based on the change of self-similarity in network traffic and normalization of network traffic. Choosing parameter is depending on attack type. Smurf attack basic parameters are logged in, guest flat login etc.

We have improved normalization method from the experimental results. The normalization is calibrated by KDDCUP dataset and our campus network normal traffic. We have used Support vector machine with KDDCUP dataset and campus network traffic in WEKA and DTREG tools. Our proposed system can detect Smurf attack in network with accuracy of 99.6%.

## 6. FUTURE WORK

In the future, we will implement and design hardware intelligence intrusion detection system using this improved normalization method and classification method.

ACKNOWLEDGEMENT

We are thankful to Asia Research Centre of National University of Mongolia for their support in the research.